\documentclass[sigconf]{acmart}
\setcopyright{none} 
\usepackage{wasysym} 

\AtBeginDocument{%
  }

\usepackage{amsmath,amsfonts}
\usepackage{graphicx}
\usepackage{textcomp}
\usepackage{xcolor}
\usepackage [english]{babel}
\usepackage [autostyle, english = american]{csquotes}
\MakeOuterQuote{"}
\usepackage{enumitem}
\usepackage{graphicx}
\usepackage{float}
\usepackage{dsfont}
\usepackage{graphics} 
\usepackage{epsfig} 
\usepackage{tabularx} 
\usepackage{multicol}
\usepackage{multirow}
\usepackage{tabularx}
\usepackage{pdflscape}
\usepackage[noindentafter]{titlesec}
\usepackage{balance} 

\usepackage{booktabs}
\usepackage{hyperref}
\usepackage{makecell}
\usepackage{pgfplots}
\pgfplotsset{width=8cm,compat=1.9} 
\graphicspath{ {images/} }
\usepackage{braket} 
\usepackage{algorithm}
\usepackage{amsthm}
\usepackage{xcolor}
\usepackage{mathtools}
\usepackage{float}
\usepackage{placeins}

\usepackage[normalem]{ulem}
\useunder{\uline}{\ul}{}
\usepackage{threeparttable}
\usepackage{makecell}
\usepackage{booktabs}

\usepackage{xspace}
\usepackage{natbib}
\usepackage{hyperref}
\usepackage{url}

\usepackage{graphicx}
\usepackage{subfigure}
\usepackage{enumitem}
\usepackage[most]{tcolorbox}
\usepackage{lipsum} 
\usepackage{changepage} 
\usepackage{fancyvrb}

\usepackage[noend]{algpseudocode}

\def\textbfmethod{{\sc \textbf{JudgeSQL}}\xspace}

\begin{document}
\urlstyle{tt}

\title{\textbfmethod: Reasoning over SQL Candidates with Weighted Consensus Tournament}

\author{Jiayuan Bai}
\authornote{Both authors contributed equally to this research.}
\email{baijiayuan@buaa.edu.cn}
\affiliation{%
  \institution{SKLSDE Lab, Beihang University}
  \city{Beijing}
  \country{China}
}

\author{Xuanguang Pan}
\authornotemark[1]
\email{panxg03@gmail.com}
\affiliation{%
  \institution{School of Software, Beihang University}
  \city{Beijing}
  \country{China}
}

\author{Chongyang Tao}
\email{chongyang@buaa.edu.cn}
\orcid{0000-0002-4162-2119}
\affiliation{%
  \institution{SKLSDE Lab, Beihang University}
  \city{Beijing}
  \country{China}
}
\author{Shuai Ma}
\email{shuaima@buaa.edu.cn}
\affiliation{%
  \institution{SKLSDE Lab, Beihang University}
  \city{Beijing}
  \country{China}
}

\renewcommand{\shortauthors}{}

\begin{abstract}
Text-to-SQL is a pivotal task that bridges natural language understanding and structured data access, yet it remains fundamentally challenging due to semantic ambiguity and complex compositional reasoning. While large language models (LLMs) have greatly advanced SQL generation though prompting, supervised finetuning and reinforced tuning, the shift toward test-time scaling exposes a new bottleneck: selecting the correct query from a diverse candidate pool. Existing selection approaches, such as self-consistency or best-of-$N$ decoding, provide only shallow signals, making them prone to inconsistent scoring, fragile reasoning chains, and a failure to capture fine-grained semantic distinctions between closely related SQL candidates.
To this end, we introduce JudgeSQL, a principled framework that redefines SQL candidate selection through structured reasoning and weighted consensus tournament mechanism. JudgeSQL develops a reasoning-based SQL judge model that distills reasoning traces with  reinforcement learning guided by verifiable rewards, enabling accurate and interpretable judgments. Building on this, a weighted consensus tournament integrates explicit reasoning preferences with implicit generator confidence, yielding selections that are both more reliable and more efficient.
Extensive experiments on the BIRD benchmark demonstrate that JudgeSQL exhibits superior SQL judgment capabilities and good cross-scale generalization and robustness to generator capacity.
\end{abstract}




\maketitle

\section{Introduction}
\label{Introduction}

Text-to-SQL has emerged as a pivotal task bridging natural language understanding and structured data access~\cite{wu2024datagptsql7b,cao2024spider2v}. By translating human language questions into executable SQL queries, Text-to-SQL enables users without database expertise to interact with complex datasets, which is critical in domains such as business intelligence, e-commerce, and data-driven content. The task is inherently challenging: it requires understanding the semantics of the query, reasoning over database schemas, and composing correct relational operations~\cite{liu2024survey}. With the rise of large language models (LLMs), there has been remarkable progress—models can now capture intricate language patterns and schema structures, generating syntactically valid SQL for diverse and complex queries, achieving state-of-the-art performance on multiple benchmarks~\cite{lei2024spider}.

Recent progress has largely focused on the \emph{generation} side, including better prompt strategies~\cite{wang2025mac}, supervised fine-tuning~\cite{pourreza2024dts}, multi-step generation, reinforcement learning~\cite{pourreza2025reasoning, ma2025sqlr1, yao2025arctic}, or generative reasoning strategies~\cite{zhang2025rewardsql,guo2025sqlforge,liu2025logiccat,liu2025dpo,pourreza2024chase}. While these approaches improve single-pass outputs, they face an intrinsic limitation: the variability and uncertainty of the sampling space, arising from the probabilistic decoding of language models and the complex, compositional reasoning required to map natural language to SQL.
Beyond enhancing individual predictions, a promising direction is Best-of-N (BoN) sampling, a widely adopted strategy for test-time scaling (TTS) in LLMs. In this approach, multiple candidate SQLs are generated—either by a single model or by an ensemble of models—and the most suitable one is selected at inference time, thereby exploring a larger hypothesis space that can enhance both accuracy and robustness.
While TTS demonstrates clear empirical benefits, it also changes the nature of the task: with a growing candidate pool, the focus naturally shifts from generating plausible queries to selecting which candidate best aligns with the intended semantics.

\begin{figure}[t]
    \centering
    \includegraphics[width=0.85\linewidth]{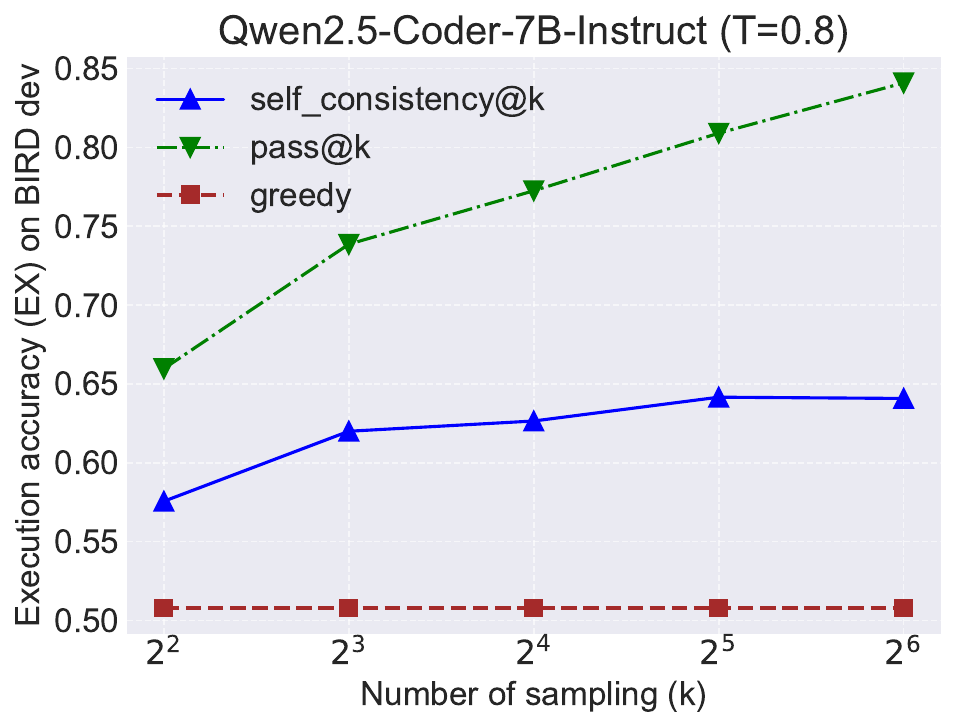}
    \vspace{-4mm}
    \caption{Trends of EX scores for greedy decoding, self-consistency, and Pass@k methods across different sampling sizes on the BIRD dev set.}
    \vspace{-5mm}
    \label{fig:trend}
\end{figure}

Existing selection strategies remain lagging behind, despite the growing complexity of the candidate space.
Current Best-of-N approaches for SQL selection typically rely on self-consistency~\cite{wang2023self}, which selects the final prediction by majority voting over execution results. While simple, such methods primarily depend on frequency and often fail to capture subtle semantic differences between candidates. More recently, a few studies have explored learned scalar-based judgment models obtained through direct fine-tuning of LLMs~\cite{pourreza2024chase}. 
While straightforward and often effective, they inherently lack mechanisms to generate intermediate reasoning steps, remain susceptible to inconsistent scoring and fragile reasoning chains, and struggle to differentiate semantically distinct queries that yield similar execution results. 
This limitation is especially pronounced in realistic scenarios, where multiple candidate SQL queries may yield comparable outputs yet differ in correctness or intent. Empirically, the performance gap between naive random choice, heuristic-based selection, and oracle selection can exceed 10 to 20 points in execution accuracy, as shown in Figure \ref{fig:trend}, demonstrating that current selection strategies leave a substantial portion of the model's potential untapped. Consequently, selecting the correct candidates has emerged as the central bottleneck in fully harnessing the reasoning capabilities of LLMs for Text-to-SQL, motivating the need for principled selection mechanisms.
 
To address these challenges, we propose JudgeSQL, a novel framework that combines structured reasoning with weighted consensus mechanisms for SQL candidate judgment.
To explore the reasoning capabilities of LLMs in SQL candidate evaluation, we first investigate the zero-shot performance of current LLMs for SQL selection by designing structured execution-guided prompts. Surprisingly, even without task-specific fine-tuning, LLMs demonstrate remarkably impressive performance compared to the widely-used SC. To further strengthen LLMs’ reasoning ability in SQL candidate judgment, we develop a reasoning-based SQL judge model that distills reasoning traces and employs reinforcement learning guided by verifiable rewards, enabling more accurate and interpretable selection of candidate queries. Finally, based on the outputs of the reasoning-based judge, we propose a weighted consensus tournament mechanism, which combines the reasoning-based judge scores with the generator’s implicit confidence to efficiently select the most reliable SQL queries.

We evaluate the proposed selection framework on the BIRD benchmark using sampled SQL candidates from several representative generation models of varying sizes and architectures. We compare both the prompted reasoning model and the reinforced reasoning model using several tournaments. The results reveal several key findings. First, RL-trained SQL judge reasoning  model boosts performance over direct prompted reasoning methods by encouraging structured reasoning chains, leading to better generalization on challenging queries. Second, by combining the selector’s explicit preferences with the generator’s implicit confidence, the weighted consensus tournament produces more reliable selections, and surpasses the traditional double round-robin tournament by achieving both higher accuracy and greater efficiency.
Moreover, performance scales positively with the number of sampled candidates, with improvements especially pronounced on the 7B model (up to 6\%). 
To summarize, our contributions are fourfold:
\begin{itemize}[leftmargin=10pt, itemsep=0pt, topsep=2pt, partopsep=0pt]
    \item We investigate the capabilities of current LLMs for zero-shot reasoning in SQL selection by designing \emph{structured execution-guided prompts} and find their performance already strong compared to the commonly used SC.
    \item We propose a \emph{reasoning-based SQL judge model} that combines reasoning trace distillation with reinforcement learning guided by verifiable rewards, resulting in more accurate and interpretable SQL candidates selection.
    \item We introduce a \emph{weighted consensus tournament mechanism} that combines reasoning-based SQL judgment and the SQL generator’s implicit confidence, achieving better performance and higher efficiency than the traditional double round-robin tournament approach.
     \item Our model achieves superior performance across various settings and demonstrates strong cross-scale generalization and robustness to generator capacity,  highlighting its potential as a reusable module for guiding SQL selection.
\end{itemize}

\vspace{-1mm}
\section{Related Works} \label{related works}
\subsection{Text-to-SQL Generation}

Early research on Text-to-SQL primarily relied on rule-based systems~\cite{popescu2004modern,li2014constructing,saha2016athena} and neural seq2seq models~\cite{basik2018dbpal,sun2018semantic,wang2020rat}, which achieved reasonable performance in restricted domains but often struggled to generalize across complex queries and heterogeneous database schemas. The advent of LLMs has significantly reshaped this landscape, offering enhanced reasoning and generalization capabilities that enable more effective handling of such complexities \citep{gao2024dail,nan2023enhancing,zhang2024benchmarking,liu2023comprehensive}. Building on these capabilities, recent studies have sought to improve the quality of SQL generation through a variety of approaches. Prompt engineering exploits carefully designed task instructions, contextual exemplars, and reasoning guidelines to guide LLMs toward semantically faithful SQL queries~\citep{pourreza2023din, gao2024dail, pourreza2024chase}. Supervised fine-tuning (SFT) with task-specific annotations further adapts models to Text-to-SQL distributions, allowing even open-source LLMs to achieve competitive performance while reducing systematic errors~\citep{pourreza2024dts, qin2025route, he2025starSQL, li2025omnisql}. In addition, multistep reasoning strategies decompose complex questions into intermediate reasoning steps or refinement phases, enabling models to progressively construct accurate SQL queries~\cite{eyal2023semantic,qu2024tasql,wang2025pi}.

To move beyond supervised paradigms, reinforcement learning  and reasoning-oriented strategies have been introduced to further enhance performance. Direct Preference Optimization-based \citep{rafailov2023dpo} methods construct pairs of preferred and dispreferred SQL candidates and are used to train models to favor outputs that are syntactically valid and semantically faithful to user intent \citep{yang2024sense, liu2025dpo, zhai2025excot}. Group Relative Policy Optimization-based \citep{shao2024deepseekmath} methods strengthen reasoning in SQL generation by refining the model's policy through sampling and comparatively evaluating multiple reasoning paths using rule-based outcome rewards. \citep{pourreza2025reasoning, ma2025sqlr1, yao2025arctic}.

\subsection{Post-processing on Text-to-SQL}
While most research focuses on SQL generation, post-processing is also crucial, as it refines queries to better match user intentions by applying strategies to enhance and correct the initial outputs.

Self-correction strategies enhance the reliability of LLM-generated SQL queries by enabling the refinement of outputs, thereby mitigating both syntactic and semantic errors. Single-refiner methods \citep{talaei2024chess,wang2025mac} incorporate execution results or self-generated feedback into the prompt context to guide a single model in refining SQL queries. Multi-agent frameworks \citep{cen2025sqlfixagent,askari2025magic,qu2025share}, employ multiple agents that collaborate to improve both syntactic and semantic correctness, placing greater emphasis on enhancing semantic accuracy through inter-agent reasoning and feedback exchange.

Complementary to self-correction, which focuses on refining individual SQL outputs post-generation, test-time scaling \citep{charlie2025tts} considers multiple candidate outputs during inference and aggregates them into a final solution to improve reliability and accuracy. A common approach within test-time scaling is output consistency \citep{wang2023self}, where models select results that remain consistent across diverse reasoning paths. Techniques such as self-consistency voting \citep{dong2023c3, gao2024dail, talaei2024chess, li2025alpha-sql} and cross-model consistency \citep{li2024pet} have been applied in text-to-SQL tasks. Since consistency alone does not ensure correctness, N-best rankers reorder outputs leveraging model-derived features or LLM-driven semantic preferences. Feature-enhanced rerankers \citep{hui2021Dynamic} exploit task-specific hidden representations and external knowledge to rank N-best SQL candidates. ZeroNL2SQL \citep{fan2024zeronl2sql} follows a similar strategy by ranking PLM-generated SQL sketches according to their semantic alignment with the input question.More recently, CHASE-SQL \citep{pourreza2024chase} and Xiyan-SQL \citep{gao2024xiyansql} adopt LLM-based judgment frameworks to rank candidate SQL queries, where CHASE-SQL performs tournament-style pairwise evaluations but lacks intermediate reasoning steps, whereas xiyan-SQL applies a listwise ranking strategy but suffers from position bias and limited robustness.

Although recent studies have demonstrated the effectiveness of test-time scaling and LLM judgment abilities, their current implementations typically rely on superficial aggregation of multiple outputs, such as consistency checks or shallow LLM-based ranking without explicit reasoning over SQL semantics or execution correctness.
To address these limitations, we explore structured execution-guided prompting for zero-shot reasoning, develop a reasoning-based SQL judge trained with verifiable reinforcement signals, and further integrate these approaches through a weighted consensus tournament that combines reasoning confidence and model reliability for robust SQL selection.

\section{Problem Formalization}
\label{sec:sampling}

In the Text-to-SQL task, given a natural language question $x \in \mathcal{X}$ and a database schema $d$, the goal is to generate a SQL query $y \in \mathcal{Y}$ such that execution result ($\mathrm{Exec}(y, d)$ correctly answers $x$. Rather than producing a single hypothesis under the model distribution \( P_\theta(y \mid x, d) \), existing approaches~\cite{talaei2024chess} approximate the space of plausible queries by sampling $N$ candidates  $\{y^{(1)}, \dots, y^{(N)}\} \sim P_\theta(y \mid x, D) $ using {temperature-scaled nucleus or top-\(k\) decoding}, typically with either powerful proprietary LLMs (e.g., GPT-4) or task-specific fine-tuned models (e.g., OminiSQL~\cite{li2025omnisql}), and then selecting among them through heuristic strategies.

While candidate generation has been widely studied, the key challenge lies in selecting the most reliable SQL from $\mathcal{Y}_x$. We therefore formulate this as learning a judgment model $\mathcal{R}(\cdot)$, interpretable as an outcome reward model, which identifies the best candidate $y^* \in \mathcal{Y}_x$ such that $\mathrm{Exec}(y^*, d)$ yields the most accurate answer to the question. To implement this, we explore comparison strategies and adopt pair-wise evaluation, which avoids the instability of set-wise scoring and enables more precise and consistent judgments.

\section{Methodology}

\begin{figure*}[t]
\centering
\includegraphics[width=\textwidth]{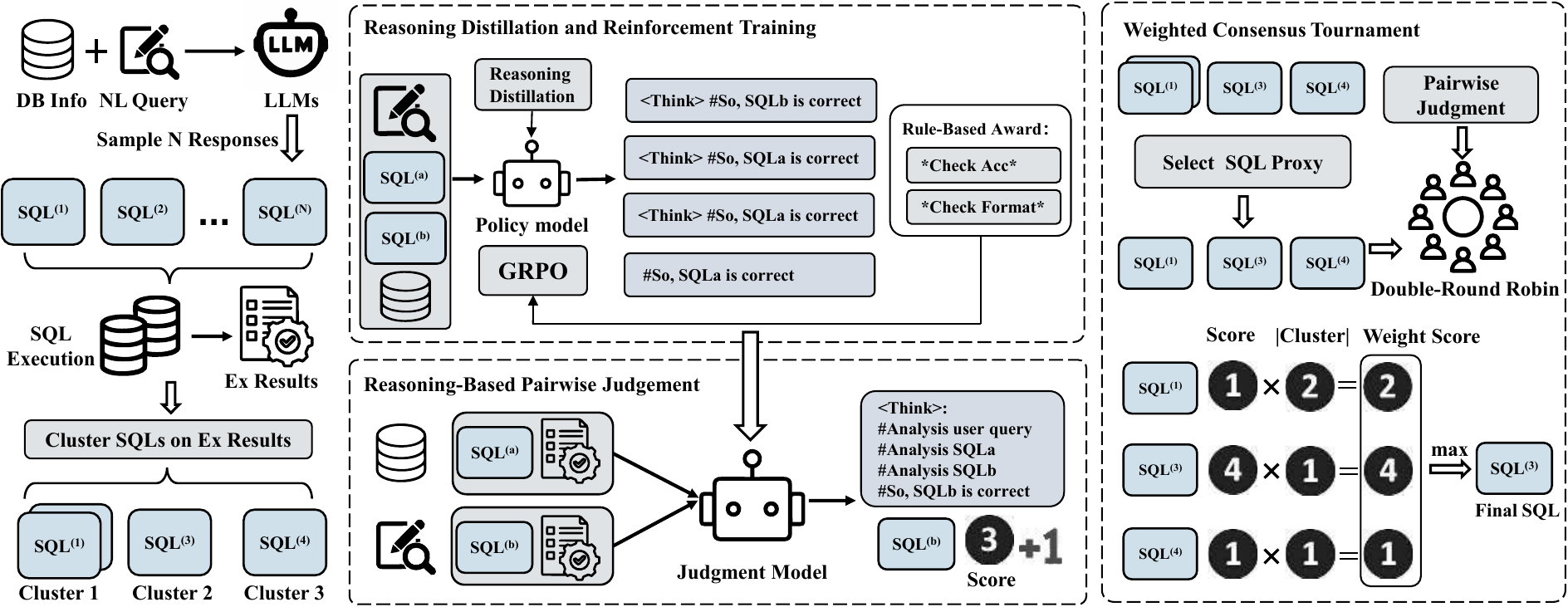}
\vspace{-6mm}
\caption{Overview of the JudgeSQL framework. }
\vspace{-4mm}
\label{fig:framework}
\end{figure*}

 \subsection{Overview of JudgeSQL}

To address the challenges of inconsistent scoring and fragile reasoning in SQL candidate selection, we propose JudgeSQL, a novel framework that integrates structured reasoning with a weighted consensus mechanism. At its core, JudgeSQL employs a reasoning-based SQL judge model that distills reasoning traces from LLMs and leverages reinforcement learning guided by verifiable rewards, enabling more accurate and interpretable evaluation of SQL candidates. Building on the judge’s outputs, a weighted consensus tournament mechanism combines the judge’s scores with the generator’s implicit confidence to efficiently select the most reliable SQL queries. An overview of our framework is illustrated in Figure~\ref{fig:framework}.
Next, we introduce our reasoning-based SQL judge model and weighted consensus tournament for SQL selection in detail.

\vspace{-1mm}
\subsection{Reasoning-reinforced SQL Judgment}
\label{sec:build_pairs}

To enhance reasoning capabilities in complex SQL judgment scenarios, we propose a reasoning-reinforced SQL judgment framework that leverages structured reasoning and verifiable feedback.
Specifically, our method employs a two-stage training strategy: reasoning trace distillation and reinforcement learning guided by verifiable rewards. In the first stage, an instruction-tuned language model (e.g., Qwen) is trained on synthetic reasoning traces for SQL candidate pairs, equipping it with structured reasoning capabilities to distinguish between semantically similar queries. In the second stage, the model is further optimized via reinforcement learning using verifiable execution-based rewards, ensuring that its judgments translate into accurate and reliable SQL selection.

\vspace{-1mm}
\paragraph{\textbf{Reasoning Distillation for
SQL Judgment.}}

Instruction-tuned language models, such as Qwen, can be adapted for SQL selection through prompting. However, without structured reasoning supervision, these models may produce inconsistent SQL choices. To address this, we introduce a distillation framework that uses small-scale synthetic SQL reasoning traces as a warm-up.

We first construct a preference dataset $\mathcal{D}_\texttt{pref}$ for SQL judgment reasoning by leveraging execution-consistent sets. For each natural language query $x$ with its ground-truth SQL $y_{gt}$ in a given training dataset (e.g., BIRD), we sample a diverse set of candidate SQLs using existing generator LLMs (e.g., Qwen Coder, OmniSQL). We execute all candidates along with the ground-truth SQL against the database and group them into execution-consistent sets $\{\mathcal{C}_1, \mathcal{C}_2, ..., \mathcal{C}_K\}$ based on their execution results. Among these $K$ sets, exactly one set $\mathcal{C}_{pos}$ contains queries that are execution-equivalent to $y_{gt}$, while the remaining sets collectively form $\mathbb{C}_{neg} = \{\mathcal{C}_1, ..., \mathcal{C}_K\} \setminus \{\mathcal{C}_{pos}\}$, representing $K-1$ distinct incorrect interpretations.
We then construct preference pairs by sampling one SQL from $\mathcal{C}_{pos}$ and pairing it with one representative SQL from a distinct negative set in $\mathbb{C}_{neg}$, yielding $N = \min(|\mathcal{C}_{pos}|, |\mathbb{C}_{neg}|)$ preference pairs. 
Formally, each pair is represented as:
\vspace{-1mm}
\begin{equation}
    \{(x, d_{uni}, y_{pos}, y_{neg}, e_{pos}, e_{neg}) | y_{pos} \in \mathcal{C}_{pos}, y_{neg} \in \mathbb{C}_{neg}\}
\end{equation}
where $e_{pos}$ and $e_{neg}$ denote the execution results of $y_{pos}$ and $y_{neg}$ respectively, and $d_{uni}$ denotes the union of the schemas involved in the two SQL candidates. To mitigate position bias, each pair is included in both orders, ensuring symmetry during training \cite{whitehouse2025j1}. 

From this preference dataset, we sample a subset
$\mathcal{D}_{\text{sub}} = \{(x^{(i)},$ $y_{pos}^{(i)}, y_{neg}^{(i)}, l^{(i)})\}_{i=1}^{M}$ where $l^{(i)}$ indicates the preferred SQL for the $i$-th input query. For each sample consisting of a SQL pair, we employ a high-capacity reasoning model $\mathcal{G}^*$ (e.g., GPT-4o~\citep{hurst2024gpt}) to generate a structured reasoning trace that explains why $y_{pos}^{(i)}$ is selected as the preferred response over $y_{neg}^{(i)}$:
\begin{equation}
\vspace{-1mm}
r^{(i)} = \mathcal{G}^*(x^{(i)},d_{uni}^{(i)}, y_{pos}^{(i)}, y_{neg}^{(i)}, l^{(i)}).
\end{equation}
\vspace{-1mm}
We filter out any samples where the oracle's reasoning is inconsistent with the known label, retaining only correctly reasoned examples. 
The final training data is constructed such that the input $\hat{x}^{(i)}$ consists of the natural language query $x^{(i)}$, {the union of the schemas involved in two SQL candidates $d_{uni}^{(i)}$}, the positive and negative SQL queries $(y_{pos}^{(i)}, y_{neg}^{(i)})$ along with their corresponding execution results $(e_{pos}^{(i)}, e_{neg}^{(i)})$, while the output is the verified reasoning trace $\hat{r}^{(i)}$ combined with the label.

\begin{equation}
\begin{aligned}
    & \hat{r}^{(i)} = r^{(i)} \oplus l^{(i)}, \quad \hat{x}^{(i)} =  (x^{(i)}, d_{uni}^{(i)}, y_{pos}^{(i)}, y_{neg}^{(i)}, e_{pos}^{(i)}, e_{neg}^{(i)}) \\
    & \mathcal{D}_{\text{distill}}  = \{\hat{x}^{(i)}, \hat{r}^{(i)} \}_{i=1}^{M}.  
\end{aligned}
\end{equation}
where $\oplus$ denotes string concatenation.
We define the SFT loss as the negative log-likelihood over the distillation dataset. To give it a more formal probabilistic interpretation, let $\hat{r}^{(i)} = (\hat{r}_1^{(i)}, \dots, \hat{r}_{T_i}^{(i)})$ denote the tokenized reasoning trace and label sequence of length $T_i$, then the objective is
\begin{equation}
\mathcal{J}_{\texttt{SFT}}(\theta) = - \sum_{i=1}^{M} \sum_{t=1}^{T_i} \log P_\theta \big( \hat{r}_t^{(i)} \,\big|\, \hat{r}_{<t}^{(i)}, \hat{x}^{(i)} \big),
\end{equation}
where $P_\theta$ represents the conditional probability assigned by the model with parameters $\theta$, and $\hat{r}_{<t}^{(i)}$ denotes the preceding tokens.  
By learning from $\mathcal{D}_{\text{distill}}$, the model captures both correct SQL selection and the corresponding reasoning path, leading to more consistent and interpretable predictions.

\paragraph{\textbf{Reinforcement Training for SQL Judgment.}} 
While knowledge distillation provides a straightforward approach to adapt a general generative model into a SQL judgment model, it can lead to overfitting on specific patterns and limit the model’s capacity to generalize its reasoning skills—an ability crucial for effective reward modeling. To overcome this limitation, we consider leveraging reinforcement learning (RL) as a more flexible and robust training paradigm that encourages reasoning-driven reward optimization. The use of RL for policy optimization is well established in the preference fine-tuning stage of LLMs, making it a natural choice for training a reasoning-based SQL jugement model. Specifically, we formulate our reward model as a policy model that learns from reasoning-informed reward signals, and we employ the GRPO algorithm to carry out reinforcement learning, formally defined as:
\begin{align*} \small
& \mathcal{J}_\texttt{GRPO} (\theta) = \mathbb{E}[\hat{x} \sim \mathcal{D}_\text{pref}, \{o_i\}_{i=1}^{|G|} \sim \pi_{\theta_{old}}(O|\hat{x})] \\
& \quad \frac{1}{|G|} \sum_{i=1}^{|G|} \left( \min\left[ \frac{\pi_{\theta}(o_i|\hat{x})}{\pi_{\theta_{old}}(o_i|\hat{x})}A_i, \text{clip}\Big(\frac{\pi_{\theta}(o_i|\hat{x})}{\pi_{\theta_{old}}(o_i|\hat{x})}, 1-\epsilon, 1+\epsilon\Big)A_i \right] \right) 
\end{align*}
Here, $\hat{x}$ denotes the formatted input sampled from the preference dataset $\mathcal{D}_\text{pref}$, and $o_i$ represents the output generated by the reward model, consisting of both the reasoning trace and the final judgment $l$. The group size $G$ indicates the number of distinct judgment traces generated for the same case. $\pi_\theta$ is the current policy model with parameters $\theta$, updated by optimizing the objective function, while $\pi_{\theta_{\text{old}}}$ denotes the fixed old policy model used to generate training samples.  
Finally, $A_i$ is the relative advantage within the group, computed by normalizing the reward $r$.

We define the reward function $r$ to jointly enforce reasoning quality and selection accuracy. A reward of one (${r}=1$) is assigned only when the model output adheres to the prescribed format—placing reasoning within <\texttt{think}>$\cdots$</\texttt{think}> and the final answer within <\texttt{answer}> $\cdots$ </\texttt{answer}>—and the predicted answer matches the ground-truth label; otherwise, the reward is zero.  Formally, this can be expressed as:
\begin{equation}
{r}(\hat{x}) = \mathbf{1}[\texttt{format\_check}(o) \wedge l_{\text{pred}} = l]    
\end{equation}
where $l_{\text{pred}}$ is the final judge answer extracted from $o$.

\subsection{Weighted Consensus Tournament (WCT)}
\label{sec:ws_tournament}
Despite a strong pair-wise judgment model, the key challenge remains efficiently selecting the correct SQL from numerous similar or execution-equivalent candidates.
Prior work~\cite{pourreza2024chase} typically relies on exhaustive pairwise comparisons, but this approach is inefficient and flawed: (i) semantically equivalent SQLs compete against each other, causing redundant comparisons and potential randomness in outcomes; (ii) each query is treated equally, ignoring the valuable signal contained in the frequency of semantically consistent candidates.

To address these limitations, we propose the \textit{weighted consensus tournament}. The key idea is to combine the robustness of set-based reasoning with the discriminative sharpness of pairwise preference. Specifically, we first form execution-consistent sets and randomly select a single representative SQL query $y^{(k)} \in \mathcal{C}_k$—to serve as the set’s proxy in subsequent comparisons\footnote{We also try selecting proxy SQL based on the highest generation likelihood from the generator LLM and achieve comparable results.}. Instead of comparing all individual queries, we let only the representatives compete in a double round-robin tournament. 
The tournament proceeds as follows: for each pair of distinct proxy SQLs $({y}^{(k)}, {y}^{(j)})$, we present both SQLs, together with their respective execution results $(e^{(k)}, e^{(j)})$ {and the union of the schemas involved in two SQL candidates $d_{uni}$}, to the reasoning judgment model $\mathcal{R}$. The model then judges which query better answers the input question $x$, defined as:
\begin{equation}
\vspace{-1mm}
    \mathcal{R}_\psi(x,d_{uni}, {y}^{(k)}, e^{(k)}, {y}^{(j)}, e^{(j)}) =
\begin{cases}
1 & \text{if } {y}^{(k)} \succ_x {y}^{(j)}, \\
0 & \text{otherwise},
\end{cases}
\end{equation}
\vspace{-1mm}
where ${y}^{(k)} \succ_x {y}^{(j)}$ indicates that ${y}^{(k)}$ is preferred over ${y}^{(j)}$ for the question $x$. 
After completing all pairwise comparisons across different execution-consistent sets, the preliminary score of each set $\mathcal{C}_k$ is computed by aggregating its wins throughout the tournament:
\vspace{-1mm}
\begin{equation}
\mathcal{S}(\mathcal{C}_k) = \sum_{j \ne k} \mathcal{R}(x,d_{uni}, {y}^{(k)}, e^{(k)}, {y}^{(j)}, e^{(j)}).
\end{equation}
\vspace{-1mm}

While the unweighted score ($\mathcal{S}(\mathcal{C}_k)$) reflects semantic preference derived from reasoning-based comparisons, it overlooks the set cardinality ($|\mathcal{C}_k|$), which encodes the generator’s implicit confidence through sampling frequency. To incorporate this prior belief, we introduce a frequency-aware weighted score:
\begin{equation}
\mathcal{S}_w(\mathcal{C}_k) = |\mathcal{C}_k| \cdot \mathcal{S}(\mathcal{C}_k), \quad k \in [1, K],
\end{equation}

This weighting scheme powerfully combines the selector's explicit preference with the generator's implicit, frequency-based confidence. The set with the highest weighted score is declared the winner, i.e., \(\mathcal{C}^* = \arg\max_k \mathcal{S}_w(\mathcal{C}_k)\). In case of a tie, we break it by selecting the set with the higher original frequency $|\mathcal{C}_k|$. Finally, we return the proxy SQL from \(\mathcal{C}^*\) as the final prediction. Algorithm~\ref{alg:WCT} presents the detailed procedure of the WCT algorithm.

Compared with a naïve double round-robin tournament, our WCT offers three major advantages. First, it reduces computational cost by comparing only a small number of proxy SQLs rather than all pairwise combinations. Second, it improves robustness by avoiding self-competition among semantically equivalent SQLs, ensuring that redundant variants reinforce rather than dilute the preference signal. Third, it exploits the consensus strength within each execution-consistent set, allowing the frequency and consistency of semantically aligned SQLs to serve as implicit confidence weights, leading to more accurate preference judgments.

\begin{algorithm}[t]
\caption{Weighted Consensus Tournament (WCT)}
\label{alg:WCT}
\begin{algorithmic}[1]
\State \textbf{Input:} Question $x \in \mathcal{X}$, Database $D \in \mathcal{D}$, Candidate set $\mathcal{Y}_x = \{y^{(1)},\dots,y^{(N)}\}$, Judgment model $\mathcal{R}_\psi$
\State \textbf{Output:} Final SQL query $y^* \in \mathcal{Y}$
\State Group candidates into execution-consistent clusters:
\Statex $\{\mathcal{C}_1,\dots,\mathcal{C}_K\} \gets \mathrm{Cluster}(\mathcal{Y}_x, D)$
\State Initialize scores: $\text{S}[k] \gets 0 \;\; \forall k \in [1,K]$
\State Select proxy SQLs for each cluster: $\{y^{(k)}\}_{k=1}^K$
\For{$k=1$ \textbf{to} $K$}
  \For{$j=1$ \textbf{to} $K$}
    \If{$k=j$} \textbf{continue} \EndIf
    \State $e^{(k)} \gets \mathrm{Exec}(y^{(k)},D), \;\; e^{(j)} \gets \mathrm{Exec}(y^{(j)},D)$
    \State $d_{uni} \gets \text{Schema}(y^{(k)})\cup \text{Schema}(y^{(j)})$
    \State $w \gets \mathcal{R}_\psi(x,d_{uni}, y^{(k)}, e^{(k)}, y^{(j)}, e^{(j)})$ \Comment{$w=1$ if $y^{(k)} \succ_x y^{(j)}$}
    \If{$w=1$}
        \State $\text{S}[k] \gets \text{S}[k] + 1$
    \Else
        \State $\text{S}[j] \gets \text{S}[j] + 1$
    \EndIf
  \EndFor
\EndFor
\State Compute weighted scores: $\text{$S_w$}[k] \gets |\mathcal{C}_k| \cdot \text{S}[k] \;\; \forall k \in [1,K]$ 
\State $\mathcal{C}^* \gets \arg\max_{k} \text{$S_w$}[k]$ 
\State $y^* \gets \text{Sample}(\mathcal{C}^*)$ 
\State \Return $y^*$
\end{algorithmic}
\end{algorithm}

\section{Experiments}

\begin{table*}[ht]
\centering
\small
\setlength{\tabcolsep}{3pt}
\caption{Evaluation results on BIRD Dev. We compare different SQL generation models and selection strategies across varying sample sizes ($N$) by each generation model. 
For reference, we also report the results of greedy sampling, which serve as a baseline for evaluating the benefits of different sampling and selection strategies.}  
\vspace{-2mm}
\resizebox{0.85\textwidth}{!}{
\begin{tabular}{lcc|cccccc}
\toprule
\multirow{2}{*}{\makecell[l]{SQL Generation\\Model}} & 
\multirow{2}{*}{\makecell[c]{Judgment\\Model}} & 
\multirow{2}{*}{\makecell[c]{Selection\\Strategy}} & 
 
\multicolumn{5}{c}{\textbf{Dev EX(\%)}} \\  
\cmidrule(l){4-9}
& & & Greedy & $N$=4 & $N$=8 & $N$=16 & $N$=32 & $N$=64 \\
\midrule
\multirow{3}{*}{\makecell[l]{Qwen2.5-Coder-\\7B-Instruct}} 
& -       & SC   &50.85& 57.63 & 62.39 & 62.71 & 63.89 & 63.89 \\

& \textsc{P}Judge & WCT   &50.85& 59.39 & 64.99 & 65.78 & 66.75 & 67.01 \\

& RJudge & WCT &50.85& 62.19 ($\uparrow$ 4.56) & 67.01 ($\uparrow$ 4.62) & 66.30 ($\uparrow$ 3.59 ) & 68.64 ($\uparrow$ 4.75) & 68.71 ($\uparrow$ 4.82) \\
\midrule
\multirow{3}{*}{\makecell[l]{XiYanSQL-7B}} 
& -    & SC   &53.59& 59.32 &62.97  & 63.89 & 64.47 & 65.06 \\

& PJudge & WCT &53.59& 61.80 & 64.99 & 65.38 & 66.75 & 68.19 \\

& RJudge & WCT  &53.59& 63.49 ($\uparrow$ 4.17) & 66.43 ($\uparrow$ 3.46) & 68.38 ($\uparrow$ 4.49) & 68.51 ($\uparrow$ 4.04) & 69.75 ($\uparrow$ 4.69) \\
\midrule
\multirow{3}{*}{\makecell[l]{OmniSQL-7B}} 
& -    & SC   &63.89& 65.51 & 66.75 & 67.28 & 67.47 & 67.28 \\

& PJudge & WCT  &63.89& 65.71 & 67.41 & 68.38 & 67.80 & 67.93 \\

& RJudge & WCT   &63.89& 68.38 ($\uparrow$ 2.87) & 69.69 ($\uparrow$ 2.94) & 70.47 ($\uparrow$ 3.19) & 70.47 ($\uparrow$ 3.00) & 70.73 ($\uparrow$ 3.45) \\
\midrule
\multirow{3}{*}{\makecell[l]{Qwen2.5-Coder-\\32B-Instruct}} 
& -    & SC   &64.47& 67.28 & 67.73  & 68.12 & 68.64 & 68.71 \\

& PJudge & WCT  &64.47& 67.86 & 68.64 & 69.17 & 69.95 & 70.01 \\

& RJudge & WCT   &64.47& 70.01 ($\uparrow$ 2.73) & 70.34 ($\uparrow$ 2.61) & 70.80 ($\uparrow$ 2.68) & 71.06 ($\uparrow$ 2.42) & 71.38 ($\uparrow$ 2.67) \\
\midrule
\multirow{4}{*}{\makecell[l]{XiYanSQL-32B}} 
& -    & SC    &63.95& 66.56 & 68.51 & 68.19 & 68.84 & 68.51 \\

& PJudge & WCT   &63.95& 67.28 & 69.23 & 69.10 & 69.81 & 70.08 \\

& RJudge & WCT  &63.95& 69.56 ($\uparrow$ 3.00) & 70.53 ($\uparrow$ 2.02) & 71.19 ($\uparrow$ 3.00) & 71.19 ($\uparrow$ 2.35) & 71.90 ($\uparrow$ 3.39) \\
\midrule
\multirow{3}{*}{\makecell[l]{OmniSQL-32B}} 
& -    & SC   &64.54& 67.14 & 67.47 & 67.21 & 67.21 & 67.14 \\

& PJudge & WCT  &64.54& 67.47 & 66.75 & 67.54 & 67.34 & 68.58 \\

& RJudge & WCT   &64.54& 69.43 ($\uparrow$ 2.29) & 69.82 ($\uparrow$ 2.35) & 70.40 ($\uparrow$ 3.19) & 70.73 ($\uparrow$ 3.52) & 70.27 ($\uparrow$ 3.13) \\
\bottomrule
\end{tabular}
}
\vspace{-3mm}
\label{tab:main_results}
\end{table*}

\subsection{Experimental Setup}
\noindent{\textbf{Datasets.}}  Following recent studies on this task \citep{talaei2024chess,pourreza2024chase,li2025alpha-sql,li2025omnisql,wang2025mac,liu2025dpo}, we conducted our experiments on BIRD~\cite{li2024can}, a large-scale cross-domain text-to-SQL benchmark that incorporates external knowledge to handle complex scenarios. BIRD contains 9,428, 1,534, and 1,789 text–SQL pairs in the training, development, and test sets, respectively.
Since the test set is not publicly available, we follow previous work and evaluate our method on the development set. We utilize the BIRD train set as the primary data source for both SFT and GRPO training. To ensure the quality of our training data, we first filter the original training set using GPT-4o~\cite{hurst2024gpt} to eliminate samples with ambiguous natural language questions or incorrect gold SQL queries. This filtering process results in a high-quality subset of 7,755 examples. Following the preference data construction procedure described in Section~\ref{sec:build_pairs}, we generate 16 candidate SQL queries for each filtered example and transform these into preference pairs, where each pair consists of execution-consistent sets containing the correct and incorrect SQL interpretations. We randomly select 200 pairs and use GPT-4o to generate their corresponding reasoning processes for the cold-start SFT stage. The remaining preference pairs are then used for GRPO training.

\vspace{1.5mm}
\noindent\textbf{Metrics.} Consisting with previous work \citep{talaei2024chess,pourreza2024chase,li2025alpha-sql,li2025omnisql,wang2025mac,liu2025dpo}, we use execution accuracy (EX) \cite{li2024can} as our evaluation metric. EX considers a generated SQL query correct if its execution result matches that of the ground-truth query, and is preferable because it avoids incorrectly penalizing valid queries when multiple correct SQL expressions exist. To evaluate a judgment model’s ability to distinguish the correct SQL, we also introduce selection accuracy (SA) in our analysis experiments.
SA is the proportion of pairwise comparisons where the model chooses the correct SQL.

\vspace{1.5mm}
\noindent\textbf{Baselines.} Our pipeline consists of two stages: SQL candidate generation and SQL selection. In the generation stage, we independently employ several representative SQL generation models, including Qwen-2.5-Coder-Instruct (7B and 32B)~\cite{hui2024qwen2}, XiYanSQL-QwenCoder (7B and 32B)~\cite{gao2024xiyansql}, and OmniSQL (7B and 32B)~\cite{li2025omnisql}, which vary in generation styles and model scales, enabling a comprehensive evaluation of our method. We compare our approach against widely used baseline selection strategies, inlcuding 1) \textbf{Self-Consistency (SC)}~\cite{wang2023self} generates multiple candidate SQL queries and selects the final prediction via majority voting over execution results, serving as a standard benchmark in Text-to-SQL~\cite{dong2023c3}; 2) \textbf{Double Round-Robin Tournament (DRT)}, introduced to Text-to-SQL by CHASE-SQL~\cite{pourreza2024chase}, evaluates all executable candidates in pairwise comparisons, where an LLM-based judgment model determines which query is more likely correct; queries accumulate scores across comparisons, and the candidate with the highest total score is chosen as the final output. 

{We also compare different judgment models under same selection strategies, including } 
1) \textbf{Prompted SQL Judgment Model (PJudge)} designs a step-by-step prompt to guide a general-purpose language model to judge candidate SQLs without task-specific training. The prompt first instructs the model to interpret the user question, then to analyze each SQL and its execution result, and finally to output a normalized decision label (A / B).
2) \textbf{Direct-SFT for SQL Judgment Models (Direct-SFT)} is trained to perform direct binary classification over candidate SQL queries without relying on any explicit reasoning process. 
\emph{Note that in this paper, we refer to our reinforced SQL judgment model as \textbf{RJudge}.}

\vspace{1.5mm}
\noindent\textbf{Implementation Details.} 
To ensure both the generality of our approach and the fairness of comparison, we employ Qwen2.5-Coder-7B-Instruct~\cite{hui2024qwen2} as the backbone SQL judgment model for all settings. The evaluation is conducted under both zero-shot and reinforced reasoning scenarios, allowing us to comprehensively examine the effectiveness and robustness of different strategies.
During the cold-start SFT stage, we perform full-parameter fine-tuning with a learning rate of 5e-6 and trained for 3 epochs. For the GRPO training stage, we utilize the VERL~\cite{sheng2025hybridflow} training framework. We maintain the learning rate at 1e-6, set  mini-batch size to 128 and train for 8 epochs. Each prompt is sampled with 8 candidate responses with a temperature of 1.0. 
The clip ratio is set to 0.2.
During inference, our pipeline consists of two stages: the SQL generation stage and the SQL tournament stage. In the SQL candidate generation stage, we employ each SQL generation model independently in our experiments, including Qwen-2.5-Coder-Instruct (7B and 32B)~\cite{hui2024qwen2}, XiYanSQL-QwenCoder (7B-2502 and 32B-2412)~\cite{gao2024xiyansql}, and OmniSQL (7B and 32B)~\cite{li2025omnisql}. 
For all generation models, we use temperature sampling (T=0.8) to generate SQL candidates.
In tournament stage for SQL selection, 
we set the decoding temperature of the judgment model to 0 and the number of samples to 1 in order to obtain deterministic outputs. The prompts for PJudge and RJudge are presented in the Appendix~\ref{sec:prompt}.

\subsection{Main Results}
Table~\ref{tab:main_results} provides a comprehensive summary of the experimental results for the proposed WCT, equipped with either prompted SQL judgment model (PJudge) or reinforced SQL judgment model (RJudge), across a variety of SQL generation models\footnote{Due to the large variation in model configurations and the limitation of computational resources, we conduct experiments on representative settings to compare the effectiveness of Direct-SFT as well as SQL candidate selection strategies.}. These include both general-purpose language models (e.g., Qwen2.5-Coder-Instruct) and domain-specific SQL generation models (e.g., XiYanSQL and OmniSQL). 
We evaluate models of different capacities (7B and 32B) across varying candidate SQL sampling sizes.

From the results, we can observe that our proposed WCT, equipped with either PJudge or RJudge, consistently outperforms SC across nearly all settings. \emph{Notably, by leveraging zero-shot reasoning in SQL selection through simple execution-guided prompts, the WCT framework already establishes a strong baseline, particularly when the sampling size is large.}
For instance, with 8 sampled SQL candidates, WCT improves EX on the Qwen2.5-Coder-7B-Instruct model from 62.39\% (SC) to 64.99\%, yielding an absolute gain of 2.6\%. 
These results highlight both the capability of current open-source models to perform reliable SQL discrimination and selection, and the effectiveness of the core WCT components in enhancing SQL selection quality.
 Furthermore, incorporating WCT with RJudge further enhances performance. Across all settings, WCT+RJudge achieves an average improvement of {3.33\%} in EX over SC and {1.35\%} over WCT+PJudge alone, demonstrating the effectiveness of the reinforced SQL reasoning judgment model. Notably, with WCT+RJudge, the 7B OmniSQL model achieves 70.73\% EX, while the 32B XiYanSQL model reaches 71.90\% EX, corresponding to improvements of {6.84\%} and {7.95\%} points over greedy sampling, respectively. This highlights the substantial potential of test-time scaling.

\begin{table}[t!]
\centering
\caption{EX (\%) under different selection strategies.
}
\vspace{-2.5mm}
\small
\setlength{\tabcolsep}{2.5pt}  
\begin{tabular}{l l|cccc}
\toprule
\multirow{2}{*}{\makecell[l]{Judgment\\Model}} & \multirow{2}{*}{\makecell[l]{Selection\\Strategy}} & \multicolumn{4}{c}{Generation Model} \\
\cmidrule(lr){3-6}
& & Qwen7B & XiyanSQL7B & Qwen32B & XiyanSQL32B \\
\midrule
\multirow{3}{*}{PJudge}
& WCT & 64.99 & 64.99 & 68.64 & 69.23 \\
& CT & 60.76 & 59.65 & 66.04 & 64.93 \\
& DRT & 63.30 & 62.91 & 67.14 & 66.69 \\

\midrule
\multirow{3}{*}{RJudge}
& WCT & 67.01 & 66.43 & 70.34 & 70.53 \\
& CT & 66.04 & 65.12 & 69.75 & 69.04 \\
& DRT & 66.88 & 65.65 & 69.75 & 68.58 \\

\bottomrule
\end{tabular}
\vspace{-4.5mm}
\label{tab:selection_strategies}
\end{table}

Across different types and sizes of generation models, our approach consistently delivers significant improvements. For example, OmniSQL-7B with WCT+RJudge achieves 70.73\% EX at 64 sampled candidates, surpassing the performance of Qwen2.5-Coder-32B-Instruct with the SC baseline (68.71\%). Similarly, XiYanSQL-32B benefits from WCT+RJudge with a +3.39\% gain over SC, indicating that large models continue to benefit from our approach without apparent saturation.
Notably, our judgment model leverages Qwen2.5-Coder-7B as its backbone and is trained on data generated by sampling SQL candidates from this model, yet it consistently improves performance across all six generation model settings, including 32B models.
This demonstrates the strong cross-scale generalization, robustness to generator capacity, and data efficiency of the model, highlighting its potential as a reusable module for guiding SQL selection across generation models of different sizes.

We further analyze the impact of sampling size on different models and selection strategies. Clearly, we can find that all generation models combined with various SQL selection strategies generally exhibit an upward trend in performance.
Notably, as the number of sampled candidates $N$ increases, the performance gains from different selection strategies are more pronounced for smaller-scale 7B generation models than for the larger 32B models.
This phenomenon is reasonable because larger models already generate high-quality SQL candidates, resulting in a strong baseline. Moreover, their candidate distributions are typically more concentrated, so additional samples often yield similar or redundant SQL queries rather than genuinely better alternatives. 
Despite this, our WCT+RJudge framework consistently outperforms SC in selecting the right SQL from large-scale models, achieving an average improvement of {2.76} points in EX score across different sampling sizes.
In addition, our WCT framework demonstrates strong computational efficiency by achieving high performance with fewer sampled candidates. For example, on OmniSQL-7B, WCT+RJudge attains 68.38\% EX using only $N=4$, surpassing the SC baseline at $N=64$ (67.28\%). This illustrates that our method can reach superior performance with significantly fewer candidates.

\begin{table}[t!]
\centering
\small
\setlength{\tabcolsep}{4pt} 
\caption{Average judgment counts per question across sampling numbers, generation models, and selection strategies.}
\vspace{-2.5mm}
\begin{tabular}{l l c c c c c}
\toprule
\multirow{2}{*}{\makecell[l]{Generation\\Model}} & \multirow{2}{*}{\makecell[l]{Selection\\Strategy}} & \multicolumn{5}{c}{Sampling Numbers} \\
\cmidrule(lr){3-7}
 & & 4 & 8 & 16 & 32 & 64 \\
\midrule
\multirow{2}{*}{Qwen7B} & DRT & 2.6 & 12.1 & 53.1 & 215.2 & 876.4 \\
 & WCT & 1.5 & 3.8 & 9.4 & 21.2 & 48.6 \\\midrule

\multirow{2}{*}{Qwen32B} & DRT & 2.0 & 9.3 & 40.1 & 167.2 & 680.9 \\
 & WCT & 0.9 & 2.0 & 3.9 & 7.3 & 15.4 \\

\bottomrule
\end{tabular}
\vspace{-3mm}
\label{tab:selection_counts_transposed}
\end{table}

\subsection{Discussions}
\paragraph{\textbf{Comparison of selection strategies.}}
To investigate the advantages of WCT, we conduct experiments comparing the execution accuracy (EX) of different selection strategies. These include DRT used in Chase-SQL~\cite{pourreza2024chase}, our WCT, and WCT without set cardinality weighting (CT). We evaluate these selection strategies combined with both PJudge and RJudge across several SQL generation models. In all experiments, the number of SQL samples is set to 8. As shown in Table~\ref{tab:selection_strategies}, WCT consistently outperforms other tournaments across almost all models and settings, confirming its advantage as a more reliable selection strategy. For PJudge, WCT achieves an improvement of +4.23 points over CT (64.99\% vs. 60.76\%) and +1.69 points over DRT (vs. 63.30\%) on Qwen-7B, with comparable gains observed on larger models (e.g., +4.30\% over CT on XiYanSQL-32B). When combined with RJudge, overall performance increases because of GRPO training. However, the relative ranking remains stable: WCT still leads with 67.01\%, followed by DRT and CT on Qwen7B.

To further underscore the advantage of WCT aginst DRT, we conduct an efficiency analysis across Qwen models at both 7B and 32B scales. Table~\ref{tab:selection_counts_transposed} reports the average number of judgments per question.
When the number of sampled candidates is 4, DRT requires, on average, 2 times more judgments than our model. At $N=64$, WCT requires only 48.6 judgments per question on Qwen-7B, compared to DRT’s 876.4 (over 18 times fewer), and 15.4 versus 680.9 (over 44 times fewer) on Qwen-32B. These results demonstrate that WCT not only achieves the highest execution accuracy but also drastically reduces selection overhead.
This efficiency stems from WCT’s grouping strategy: while DRT compares all candidates exhaustively, WCT clusters SQLs based on their execution results and performs pairwise comparisons only between different groups, substantially reducing the total number of comparisons.

\begin{table}[t!]
\centering
\setlength{\tabcolsep}{2.1pt}
\caption{Execution Accuracy (EX) and Selection Accuracy (SA) (\%) across generation models under different training recipes. 
}
\vspace{-2mm}
\small
\resizebox{0.99\linewidth}{!}{
\begin{tabular}{l|cccccccc}
\toprule
\multirow{3}{*}{\makecell[l]{Training \\Recipe}}
& \multicolumn{8}{c}{Generation Model} \\ 
\cmidrule(l){2-9}
& \multicolumn{2}{c}{Qwen-7B} 
& \multicolumn{2}{c}{XiyanSQL-7B} 
& \multicolumn{2}{c}{Qwen-32B} 
& \multicolumn{2}{c}{XiyanSQL-32B} \\
\cmidrule(lr){2-3} \cmidrule(lr){4-5} \cmidrule(lr){6-7} \cmidrule(l){8-9}
& EX & SA & EX & SA & EX & SA & EX & SA \\
\midrule
Base & 64.99 & 71.33 & 64.99 & 73.91 & 68.64 & 62.91 & 69.23 & 67.53 \\
Direct-SFT & 66.30 & 80.13 & 65.97 & 81.38 & 69.30 & 71.38 & 69.56 & 73.57 \\
GRPO & 66.23 & 78.97 & 66.88 & 79.97 & 69.56 & 72.34 & 70.21 & 73.14 \\ \midrule
RJudge  & 67.01 & 82.55 & 66.43 & 83.04 & 70.34 & 75.92 & 70.53 & 75.51 \\

\bottomrule
\end{tabular}
}
\vspace{-1mm}
\label{tab:training_recipes_combined}
\end{table}

\paragraph{\textbf{Effects of training recipes.}}
To better understand the impact of model training on judgment performance, we compare supervised fine-tuning with only judgment label prediction (Direct-SFT) and GRPO training without distillation warmup (GRPO) under the WCT selection strategy across different SQL generation models, using $N=8$ sampled SQL candidates per question.
Since the final outcome of the tournament does not always fully reflect the capability of the judgment model, we also report the judgment selection accuracy.
As shown in Tables~\ref{tab:training_recipes_combined}, performance improves progressively with recipe refinement.
SFT with only judgment label prediction yields clear gains over the base model, particularly in selection accuracy, improving by 6.04-8.80\% across different generation models. 
GRPO alone yields mixed results and may underperform compared to SFT, but when combined with a warmup—starting with SFT distillation from GPT-4o followed by GRPO—it consistently achieves the best performance (e.g., 67.01\% execution accuracy and 82.55\% selection accuracy on Qwen-7B). The results demonstrate that warmup with SFT distillation can effectively facilitate GRPO training for SQL judgment, enhancing the model’s SQL reasoning and judgment capabilities.
{Interestingly, we observe that selection accuracy varies with the generation model, with 7B model being generally easier to discriminate than those from 32B model. This suggests that the quality and diversity of the generated candidates can significantly impact the judgment model's performance.}

\begin{table}[t]
\centering
\small
\setlength{\tabcolsep}{4pt}
\caption{Comparison of different judgment models under zero-shot reasoning with SQL generation(Dev EX\%).}
\vspace{-1mm}
\begin{tabular}{l|cccc}
\toprule
\multirow{2}{*}{\makecell[l]{Judgment\\Model}} & \multicolumn{4}{c}{Generation Model} \\
\cmidrule(lr){2-5}
                              & Qwen7B & XiyanSQL7B & Qwen32B & XiyanSQL32B \\
\midrule
SC &62.39&62.97&67.73&68.51 \\
\midrule
Deepseek-6.7B& 64.02& 65.45 & 68.64 & 68.97\\
Qwen-7B  &64.99 & 64.99 & 68.64 & 69.23  \\
Llama-8B &64.34 & 64.47 & 68.12 & 68.51 \\
\bottomrule
\end{tabular}
\vspace{-2mm}
\label{tab:judge_models_zeroshot}
\end{table}

\paragraph{\textbf{Comparison of different judgment models}}
While our primary experiments employed Qwen-7B as the base judgment model, it demonstrates strong robustness and generalization across models of different sizes and types. We are also interested in evaluating the generalizability of WCT across diverse judgment model architectures, including Llama and DeepSeek.
This extension aims to verify that the performance gains observed are attributable to our proposed method itself rather than specific characteristics of the Qwen model. Table~\ref{tab:judge_models_zeroshot} reports the performance of Llama3-8B-Instruct~\citep{dubey2024llama} and Deepseek-coder-6.7B-Instruct~\citep{guo2024deepseek} as zero-shot judgment models across different SQL generation models, using a sampling size of 8.
We can find that all judgment models generally outperform the baseline SC method.
Among the three models, Qwen and Deepseek achieve better performance, while Llama shows slightly worse results, likely due to its relatively weaker code generation and reasoning capabilities, as noted in previous studies.
Despite these differences, we can still conclude that the WCT framework is robust, operating stably across various underlying judgment models.

\begin{figure}[t!]
\centering
\includegraphics[width=0.995\linewidth]{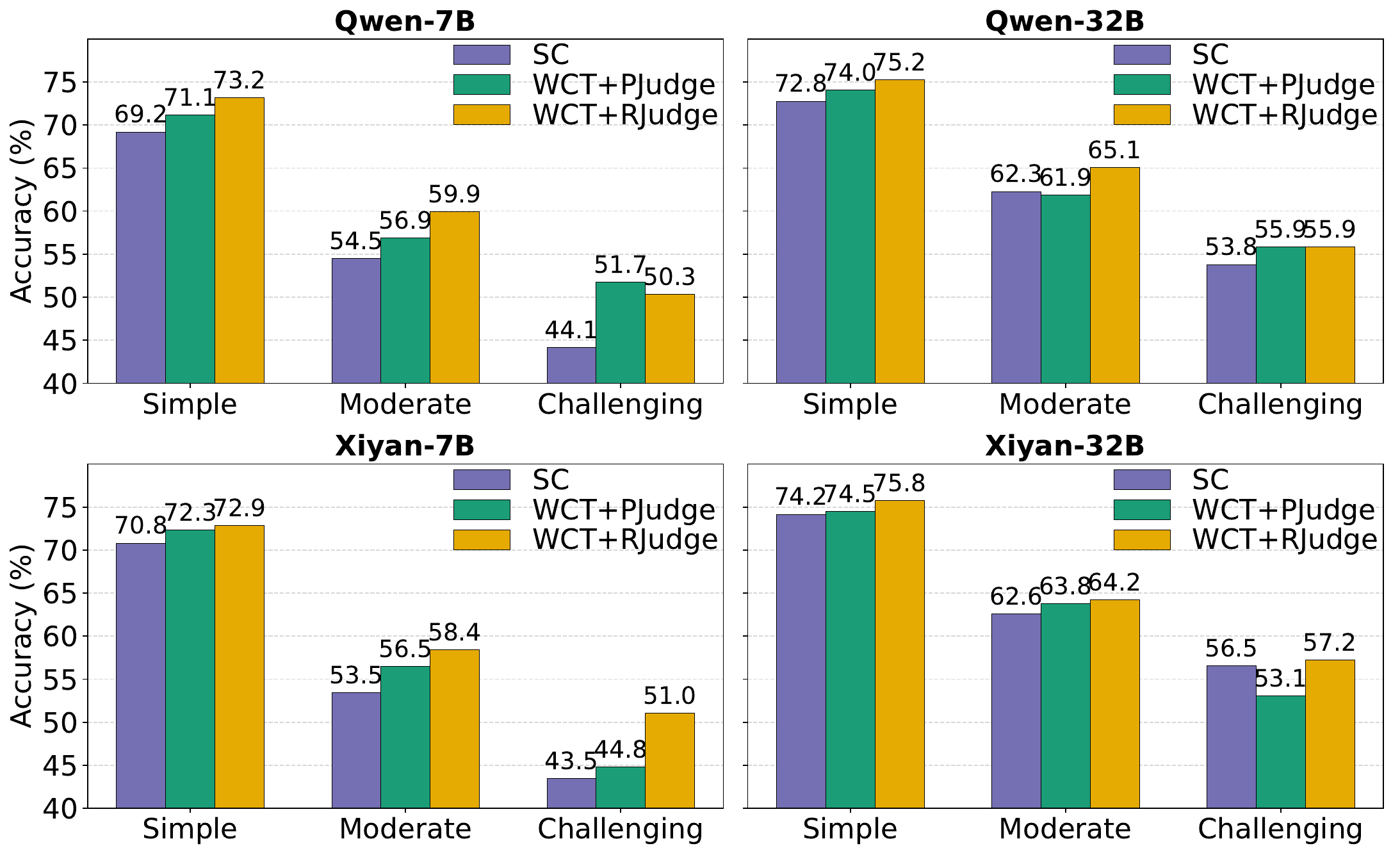}
\vspace{-4mm}
\caption{Performance by query difficulty on BIRD dev set. }
\vspace{-3mm}
\label{fig:difficulty_analysis}
\end{figure}

\paragraph{\textbf{Performance on queries of varying difficulty}}
{To better understand the performance of our approach across varying levels of query complexity, we analyze the results of different methods on the BIRD development set, as shown in Figure~\ref{fig:difficulty_analysis}. Three difficulty levels—Simple, Moderate, and Challenging—are annotated by human experts in the BIRD benchmark.
The WCT+PJudge and WCT+RJudge methods demonstrate consistent superiority over the SC baseline across most of the difficulty levels. On simple level queries, WCT+RJudge improves accuracy by +1.6\% to +4.00\% (e.g., Qwen-7B: 69.2\% → 73.2\%; Xiyan-32B: 74.2\% → 75.8\%). For moderate queries, the gains are more substantial, ranging from +1.6\% to +5.4\% (e.g., Qwen-7B: 54.5\% → 59.9\%; Qwen-32B: 62.3\% → 65.1\%). Most notably, for challenging queries where precision is most critical, WCT+RJudge provides the most valuable improvements. The absolute gains reach +7.5\% (Qwen-7B: 44.1\% → 50.3\%, +6.2\%; Xiyan-7B: 43.5\% → 51.0\%, +7.5\%). This is particularly significant given the already low baseline performance on complex queries. This consistent advantage of our execution-aware selection methods confirms their benefit regardless of query complexity, with particularly valuable gains on challenging queries where every percentage point matters for practical deployment}

\section{Conclusion}
In this paper, we introduce JudgeSQL, a framework for SQL candidates selection that integrates structured reasoning and consensus-based decision mechanisms. JudgeSQL enhances judgment accuracy and interpretability through reinforcement learning with verifiable rewards, alleviating the limitations of heuristic strategies such as execution-based majority voting and prior pairwise judgment methods. The weighted consensus tournament additionally combines reasoning-based confidence with generator likelihood, enabling more reliable and efficient SQL selection. Experiments on BIRD show that JudgeSQL consistently outperforms existing methods, with good  generalization and robustness to generator capacity.



\bibliographystyle{ACM-Reference-Format}
\balance
\bibliography{custom}

\clearpage
\appendix
\appendix
\section{Appendix}

\subsection{Case Study}
While the quantitative results above demonstrate the superiority of GRPO over the base model, we further present a case study to illustrate how GRPO enhances the reasoning process by producing more accurate and reliable reasoning thought. Specifically, Table~\ref{tab:GRPOinfer} presents the inference result for question ID 506, which is annotated as a challenging case in the BIRD development set. Notably, this example is correctly solved by the RSJ model, whereas both the PSJ model and the SFT model fail on it.
This case study tasked the judge model with selecting the correct SQL query to find “the percentage of card set with Japanese translations that are non-foil only.” The model’s reasoning demonstrated a sophisticated grasp of the problem. It first identified that the core operation must be performed at the set level, not the card level. This allowed it to diagnose the flaw in Candidate A, an unnecessary join with the cards table, which it reasoned would cause a Cartesian product and inflate the result to an invalid 153.9\%. Conversely, the model validated Candidate B for its correct use of a subquery to filter and aggregate at the series level, which aligned perfectly with the task's semantics. This shows the model's ability to map natural language to database operations and perform logical code review grounded in systematic reasoning.\\
\newtcolorbox{casestudy}[1][]{
  colback=blue!10,        
  colframe=blue!80!black, 
  fonttitle=\bfseries,       
  title=Case Study,          
  #1                         
}

\begin{figure}[H]
\centering
\begin{minipage}{0.99\linewidth}
\begin{casestudy}[title=Case Study: Task description]
\textbf{Question:} For all the set of cards that has Japanese translation, what is the percentage of them are only available in non-foil?

\textbf{Candidate A SQL:}
\begin{verbatim}
SELECT CAST(SUM(CASE WHEN T2.isNonFoilOnly = 1 
       THEN 1 ELSE 0 END) AS REAL) * 100
/ SUM(T1.language = 'Japanese')
FROM set_translations AS T1
INNER JOIN sets AS T2 ON T1.setCode = T2.code
INNER JOIN cards AS T3 ON T2.code = T3.setCode;
\end{verbatim}

\textbf{Execution Result of A:} [[153.8992408557626]]

\textbf{Candidate B SQL:}
\begin{verbatim}
SELECT SUM(CASE WHEN isNonFoilOnly = 1 
       THEN 1 ELSE 0 END) * 100.0 / COUNT(*) 
       AS percentage_non_foil
FROM sets
WHERE code IN (SELECT setCode FROM set_translations 
      WHERE language = 'Japanese');
\end{verbatim}

\textbf{Execution Result of B:} [[11.570247933884298]]

\vspace{1em}

The model will be prompt to think step by step and choose the candidates that correctly answer the question 

\end{casestudy}
\end{minipage}
\caption{The task description of the example case study.}
\label{tab:case}
\end{figure}

\vspace{1em}

\begin{figure}[H]
\centering
\begin{minipage}{0.99\linewidth}
\begin{casestudy}[title=GRPO model response]

\textbf{Reasoning:} To determine the correct SQL query, we need to carefully analyze the given schema and the natural language question. The question asks for the percentage of sets that have a Japanese translation and are only available in non-foil format.

1. We need to join the \texttt{set\_translations} table with the \texttt{sets} table on \texttt{setCode} to filter sets that have a Japanese translation.

2. We need to calculate the percentage of these sets that are only available in non-foil format.

Let's analyze the two candidate SQL queries:

\textbf{Execution Result of B:} [[11.570247933884298]]

Both queries involve joining the \texttt{set\_translations} table with the \texttt{sets} table to filter sets with a Japanese translation and then calculating the percentage of non-foil sets among them.

Candidate B filters sets with a Japanese translation and calculates the percentage of non-foil sets correctly. Candidate A seems to have an issue with the division and filtering logic.

Given the execution results, Candidate B returns a reasonable percentage, while Candidate A returns an unusually high value.

Therefore, the correct query is Candidate B.

\end{casestudy}
\end{minipage}
\caption{The inference process of GRPO model for this case study.}
\label{tab:GRPOinfer}
\end{figure}
\clearpage

\subsection{Prompt Template for RJudge and PJudge Model}
\label{sec:prompt}
\begin{center}
\begin{minipage}{0.95\textwidth}
\begin{casestudy}[title=Prompt for PJudge model]
You first thinks about the reasoning process in the mind and then provides the user with the answer.\\

Task Overview:
You are a data science expert. Below, you are provided with a database schema, a natural language question, two candidates SQL and its corresponding execution result. Your task is to understand the schema and choose the correct SQL which answers the natural language question from the two candidates.\\

Database Engine:\\
SQLite\\

Database Schema:\\
\{DATABASE\_SCHEMA\}\\

This schema describes the database's structure, including tables, columns, primary keys, foreign keys, and any relevant relationships or constraints.

Question:\\
\{EVIDENCE\}
\{QUESTION\}\\

Here are two candidate SQLs and their execute results:\\
Candidate A: 
\{CANDIDATE\_A\_QUERY\}\\
Execution result of A: 
\{CANDIDATE\_A\_RESULT\}\\

Candidate B: 
\{CANDIDATE\_B\_QUERY\}\\
Execution result of B: 
\{CANDIDATE\_B\_RESULT\}\\

Instructions:\\
- Before choosing the final answer, please think through the steps of how to confirm its correctness. You should think through these steps:\\
1. Understanding the user question requirements:  – What information is being requested? What are the key columns and filters?\\
2. Analyzing the Database schema: – Which tables and joins are necessary to answer the question? What are the key relationships?\\
3. Evaluating candidate A and its execution results: – Does it select the correct columns? Does it apply the correct conditions and joins? Does the result match expectations?\\
4. Evaluating candidate B and its execution results: – Does it select the correct columns? Does it apply the correct conditions and joins? Does the result match expectations?\\
5. Comparing two candidates and their execution results: -– Identify differences between the two queries in terms of structure, logic, and execution result.\\
6. Determining the Correct answer: – Select the query that best satisfies the question requirements, with correct columns, filters, and joins.\\
 
Remember:\\
- The correct SQL should return all of the information asked in the question without any missing or extra information. If the question asks for a specific column, the correct candidate SQL only include that column in the SELECT clause, nothing more, nothing less.\\
- The empty execution result "[]" does not necessarily mean that the SQL query is incorrect; it might simply indicate that the database does not contain such data. \\

Output Format:
Show your work in <think> </think> tags. And return your answer 'A' or 'B' in <answer> </answer> tags. For example, <think>reasoning process here</think><answer>A</answer> if the candidate A is correct, <think>reasoning process here</think><answer>B</answer> if the candidate B is correct.
\end{casestudy}
\end{minipage}
\end{center}

\begin{figure*}[h]
\centering
\begin{minipage}{0.95\textwidth}
\begin{casestudy}[title=Prompt for RJudge model]
You first thinks about the reasoning process in the mind and then provides the user with the answer.\\

Task Overview:
You are a data science expert. Below, you are provided with a database schema, a natural language question, two candidates SQL and its corresponding execution result. Your task is to understand the schema and choose the correct SQL which answers the natural language question from the two candidates.\\

Database Engine:\\
SQLite\\

Database Schema:\\
\{DATABASE\_SCHEMA\}\\

This schema describes the database's structure, including tables, columns, primary keys, foreign keys, and any relevant relationships or constraints.\\

Question:\\
\{EVIDENCE\}
\{QUESTION\}\\

Here are two candidate SQLs and their execute results:\\
Candidate A: 
\{CANDIDATE\_A\_QUERY\}\\
Execution result: 
\{CANDIDATE\_A\_RESULT\}\\

Candidate B: 
\{CANDIDATE\_B\_QUERY\}\\
Execution result: 
\{CANDIDATE\_B\_RESULT\}\\

Instructions:\\
- The correct SQL should return all of the information asked in the question without any missing or extra information. If the question asks for a specific column, the correct candidate SQL only include that column in the SELECT clause, nothing more.\\
- The empty execution result "[]" does not necessarily mean that the SQL query is incorrect; it might simply indicate that the database does not contain such data.\\
- Before choosing the final answer, please think through the steps of how to confirm its correctness.\\

Output Format:
Show your work in <think> </think> tags. And return your answer 'A' or 'B' in <answer> </answer> tags. For example, <think>reasoning process here</think><answer>A</answer> if the candidate A is correct, <think>reasoning process here</think><answer>B</answer> if the candidate B is correct.

\end{casestudy}
\end{minipage}
\end{figure*}

\end{document}